\documentclass[a4paper]{article}

\usepackage{INTERSPEECH_v2}
\usepackage{graphicx}
\usepackage{amsmath}
\usepackage{textcomp, xspace}
\usepackage{multirow}

\title{Towards Zero-Shot Frame Semantic Parsing for Domain Scaling}
\name{Ankur Bapna, Gokhan T\"{u}r, Dilek Hakkani-T\"{u}r, Larry Heck}
\address{Google Research, Mountain View}
\email{\{ankurbpn,gokhant,dilekh,larryheck\}@google.com}

\begin{document}
\newcommand\la{\textlangle\xspace}  
\newcommand\ra{\textrangle\xspace}
\maketitle
\begin{abstract}
State-of-the-art slot filling models for goal-oriented human/machine
conversational language understanding systems rely on deep learning
methods. While multi-task training of such models alleviates the need
for large in-domain annotated datasets, bootstrapping a semantic parsing
model for a new domain using only the semantic frame, such as the
back-end API or knowledge graph schema, is still one of the holy grail
tasks of language understanding for dialogue systems. This paper proposes a deep learning
based approach that can utilize {\em only the slot description}
in context without the need for any labeled or unlabeled in-domain
examples, to quickly bootstrap a new domain. The main idea of this paper is to leverage
the encoding of the slot names and descriptions within a multi-task
deep learned slot filling model, to implicitly align slots across domains. The proposed approach is
promising for solving the domain scaling problem and eliminating the
need for any manually annotated data or explicit schema alignment. Furthermore, our 
experiments on multiple domains show that this approach
results in significantly better slot-filling performance when
compared to using only in-domain data, especially in the low data regime.
\end{abstract}

\section{Introduction}

In traditional goal-oriented dialogue systems, user utterances are
typically understood in terms of hand-designed semantic frames
comprised of domains, intents and slots~\cite{SLUBook}. Understanding
the user utterance involves (i) detecting the domain of the utterance,
(ii) classifying the intent of the utterance based on the semantic frame
corresponding to the detected domain and (iii) identifying the values or sequence of tokens corresponding to each slot
in the semantic frame. An example semantic frame is shown in
Figure~\ref{fig:ex} for a flight related query: {\em find flights to
  new york tomorrow}. \\
Most modern approaches for conversational language understanding
involve training machine learning models on annotated training
data~\cite[among others]{young,luna,WangDengAcero05}. Deep learning models typically outperform most other approaches in the domain of large scale
supervised learning and this has been shown to be the case for spoken
language understanding~\cite[among
  others]{RNN-TASL,dilekIS16,Kurata:emnlp16,ruhi-dbn:2011}. However,
despite recent advancements and tremendous research activity in
semi-supervised and unsupervised learning, these models still require
massive amounts of labeled data to train. \\
In recent years, motivated by commercial applications like Apple Siri,
Microsoft Cortana, Amazon Alexa, or Google Assistant, there is
significant interest in enabling users to add more functionality and power to their respective assistants. However, although these assistants can handle queries corresponding to certain narrow domains with high reliability, the ability to understand user queries across a wide range of domains, in a robust manner, is still missing. This is a significant bottleneck that restricts the ability to crowd-source the addition of new actions or skills to these assistants. \\
With recent advances in deep learning, there is renewed excitement
around latent semantic representations which can be trained for a
multitude of covered domains via transfer learning. An earlier
work proposed training a single multi-task deep learning
model covering all domains, providing implicit shared feature learning
across domains~\cite{dilekIS16}. The approach showed 
significantly better overall semantic template level performance. Similar experiments on using shared feature extraction layers for slot-filling across
several domains have demonstrated significant performance improvements
relative to single-domain baselines, especially in low data regimes~\cite{aaronIS16}. \\
\begin{figure}[t]
  \centering
  \begin{tabular}{|ccccccc|}
    \hline
    $W$&find&flights&to&new&york&tomorrow\\
    &$\downarrow$&$\downarrow$&$\downarrow$&$\downarrow$&$\downarrow$&$\downarrow$\\
    $S$&O&O&O&B-Dest&I-Dest&B-Date\\
    $D$&\multicolumn{2}{l}{flight}&&&&\\
    $I$&\multicolumn{2}{l}{find\_flight}&&&&\\
    \hline
  \end{tabular}
  \caption[Figure]{An example semantic parse of an utterance ($W$) with slot ($S$), domain ($D$), intent ($I$) annotations, following the IOB (in-out-begin) representation for slot values.}
  \label{fig:ex}
\vspace{-1.0cm}
\end{figure}
\begin{table*}[t]
  \begin{small}
  \begin{center}
  \caption{Sample utterances from each domain}
  \label{tab:1}
  \begin{tabular}{|p{2.5cm}||p{9cm}||p{1.5cm}|}
  \hline
    Domain & Sample & \# Samples\\
    \hline
    bus\_tickets & I need 2 adult and 6 senior bus tickets from St . Petersburg to Concord. & 500\\
    \hline
    book\_room &I need a hotel room for 5 guests to check-in next Friday & 500\\
    \hline
    flights\_1 & book a flight to logan airport 3 / 23 to jan 2 & 10000\\
    \hline
    flights\_2 & Search for flights to Philly one - way with promo code 54ZFHK33 & 500\\
    \hline
    fare & How much is it on Lyft to go from Saratoga to Fremont & 1000\\
    \hline
    find\_restaurants & chinese places to eat that are not expensive & 1000\\
    \hline
    appointments & set up a patient follow - up with ProHealth Chiropractic & 2000\\
    \hline
    reserve\_restaurant & I need a table at Sun Penang on December 24th & 5000\\
    \hline
    book\_cab& book a ride to 9192 johnson street with uber for 6 passengers & 2000\\
    \hline
    book\_hotel & book me a hotel room in Cincinnati that costs less than \$300 & 1000\\
  \hline
  \end{tabular}
  \end{center}
  \end{small}
\vspace{-0.5cm}
\end{table*}
In this study we explore semi-supervised slot-filling based on deep
learning based approaches that can utilize natural language {\em slot label
  descriptions}. This alleviates the need for large amounts of labeled or unlabeled
in-domain examples or explicit schema alignment, enabling developers to quickly bootstrap new domains. Similar ideas have previously been shown to work 
for domain classification, where domain names were leveraged to generate representations in a shared space with query representations ~\cite{Yann-ICLR}. Applying this idea to a full semantic frame is
much more complex and requires building a general slot or concept
tagger over a large set of domains. \\
The architecture for the general concept tagger is similar to those proposed in recent Question Answering (QA) and Machine Reading literature. We build upon the idea of using a learned question encoding and using it for extractive question answering from input passages \cite{seqlabQA,lee2016learning,machineread}. However, our proposed use case involves using slot encodings learned from their natural language descriptions and training the model on a multitude of small in-domain datasets to generalize to new domains, instead of training and evaluating on larger open-domain QA datasets. \\
Through our experiments we demonstrate that our model learns to identify slots across domains, from small amounts of training data, without the need for any explicit schema alignments. Such an approach can
significantly alleviate the domain scaling problem and reduce the need for
additional manually annotated data when bringing up a new domain. \\
In Section~\ref{sec:task} we describe the task and the dataset,
followed by descriptions of the baseline model, the multi-task model
and the concept tagger in Section~\ref{sec:model}. This is followed by
experimental results in Section~\ref{sec:exp} and discussion in Section~\ref{sec:disc}.

\section{Slot Filling}
\label{sec:task}

In most spoken dialogue systems, the semantic structure of an application domain is defined in terms of \textit{semantic frames}. Each semantic frame contains several typed components called ``\textit{slots}.'' For the example in
Figure~\ref{fig:ex}, the domain {\em Flights} may contain slots
like {\em Departure\_City}, {\em Arrival\_City}, {\em
  Departure\_Date}, {\em Airline\_Name}, etc. The task of slot filling
is then to instantiate slots in semantic frames from a given user query or utterance. \\
More formally, the task is to estimate the sequence of tags $Y = y_1, ..., y_n$ in
the form of IOB labels as in~\cite{raymond-riccardi07} (with 3 outputs
corresponding to `B', `I' and `O'), and as shown in
Figure~\ref{fig:ex} corresponding to an input sequence of tokens $X =
x_1, ..., x_n$.

\begin{table*}[t]
  \begin{small}
  \begin{center}
  \caption{Slot schema / descriptions used for the concept tagger for each domain}
  \label{tab:2}
  \begin{tabular}{|p{2.5cm}||p{13.5cm}|}
  \hline
    Domain & Slot descriptions \\
    \hline
    bus\_tickets & departure time, number of adult passengers, arrival location, number of child passengers, number of senior passengers, departure location, promotion code, date of departure, trip type, discount type, date of return\\
    \hline
    book\_room & features, property type, number of beds, number of guests, maximum price per day, location, check out date, room type, check in date \\
    \hline
    flights\_1 & flight class, date of second departure, number of passengers, second from location, flight type, from location, search type, departure date, second to location, to location, non stop, return date \\
    \hline
    flights\_2 & origin, \# seniors, departure date, \# adults, destination, return date, price type, promotion code, trip type \\
    \hline
    fare & origin, destination, transit operator \\
    \hline
    find\_restaurants & amenities, hours, neighborhood, cuisine, price range \\
    \hline
    appointments & services, appointment time, appointment date, title \\
    \hline
    reserve\_restaurant & number of people, restaurant name,reservation date, location, cuisine, restaurant distance, reservation time, meal, price range, rating \\
    \hline
   book\_cab & pickup location, drop off location, \# passengers, cab operator, type of cab, departure time\\
    \hline
    book\_hotel & amenities, departure date, arrival date, price range, \# rooms, hotel name, ratings, room type, location, duration of stay \\
  \hline
  \end{tabular}
  \end{center}
  \end{small}
\vspace{-0.5cm}
\end{table*}

\begin{figure}
  \centering
	\includegraphics[width=0.48\textwidth]{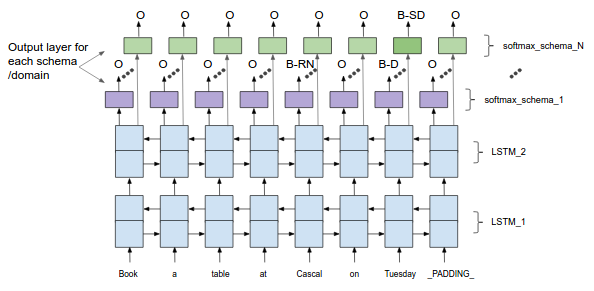}
	\caption{Multi-task stacked LSTM architecture}
	\label{fig:multi}
\vspace{-0.5cm}
\end{figure}

In literature,
researchers usually employ known sequence labeling methods for filling
frame slots of an application domain using a labeled training data
set. With advances in deep learning, the best performing academic
slot filling systems rely on recurrent neural network (RNN) or Long Short Term Memory (LSTM) based
models. RNNs were first used for slot filling by Yao {\em et
  al.}~\cite{yao2013RNN} and Mesnil {\em et al.}~\cite{mesnil2013RNN}.
A comprehensive compilation of RNN based slot filling
approaches was described by Mesnil {\em et al.}~\cite{RNN-TASL}. State-of-the-art slot filling methods usually rely on bidirectional
LSTM models~\cite[among
  others]{dilekIS16,RNN-TASL,Kurata:emnlp16,vu2016bi,vukoticIS16}. Extensions
include encoder-decoder models\cite[among others]{bingIS16,zhu2016} or
memory networks~\cite{vivianIS16}. \\
For this study we crowd sourced natural language text data for 10 domains. The schema corresponding to each of
these domains is described in Table~\ref{tab:2}.
One noticeable feature of our datasets is a lack of fine-grained slot types as compared to the popular ATIS dataset ~\cite{price1990evaluation} which contains over a hundred distinct slots. \\
For the collection, a list of possible
slot-value combinations was generated from the Google knowledge graph
manually, and used to prompt crowd-workers with slot-value pairs. The crowd-workers were instructed to ask their digital assistant to complete certain tasks with given slot-value based arguments. The collected utterances were then either automatically labeled if a verbatim match was found for the slot-values, or sent out to a second set of raters for labeling. For the labeling job the crowd workers were instructed to label spans corresponding to slot values in the instantiated samples.\\
 Table~\ref{tab:1} shows the list of these domains with
representative example queries and the total number of training samples available. Test sets were constructed using the same framework. 
All the collected data was tokenized using a standard tokenizer and lower-cased before use since capitalization was seen to be indicative of slot values. All digits were replaced with special ``\#" tokens following ~\cite{aaronIS16}.

\section{Models}
\label{sec:model}
In this study we explore the idea of zero-shot slot-filling, by
implicitly linking slot representations across domains by using the
label descriptions of the slots. We compare the performance of three model architectures on varying amounts of training data:
\vspace{-0.1cm}
\begin{itemize}
\item Single task bi-directional LSTM \vspace{-0.1cm} 
\item Multi-task bi-directional stacked LSTM model\cite{dilekIS16,aaronIS16} \vspace{-0.1cm}
\item Concept tagging model using slot label descriptions \vspace{-0.05cm}
\end{itemize}
For all our experiments we use 200 dimensional word2vec embeddings trained
on the GNews corpus ~\cite{word2vec}. Tokens not present in the pre-trained embeddings were replaced by a \_OOV\_ token. Each model was
trained for 50000 steps using the RMSProp optimizer and tuned on the dev set performance before evaluation on the test set. \\
For evaluation, we compute the token F1 for each slot independently and report the weighted average over all slots for the target domain. We use token F1 instead of the traditional slot F1 since token level evaluation
results in softer penalization for mistakes, to mitigate span inconsistencies in
the crowd sourced labels. 

\subsection{Baseline single task model}

We use a single domain bidirectional LSTM as our baseline model. The
model consists of the embedding layer followed by a 128 dimensional
(64 dimensions in each direction) bidirectional LSTM. This is followed
by a softmax layer that acts on the LSTM state for every token to
predict the IOB label corresponding to the token.

\subsection{Multi-task model}

The multi-task model consists of 2 stacked bidirectional LSTM layers
with 256 dimensions each (128 dimensions in each direction). Both LSTM
layers are shared across all domains, followed by domain specific
softmax layers, following~\cite{aaronIS16}. The model was trained using a batch size of 100 with alternating batches from different domains. To avoid over-training the model on the larger domains, the number of batches chosen from each domain was proportional to the logarithm of the number of training samples from the domain.
The conceptual model architecture is depicted in
Figure~\ref{fig:multi}.

\subsection{Zero-Shot Concept Tagging Model}

The main idea behind the zero-shot concept tagger is to leverage the slot names
or descriptions in a domain-agnostic slot tagging model. Assuming that the slot description is semantically accurate, if one of the already covered domains contains a similar slot, a continuous representation of the slot obtained from shared pre-trained embeddings can be leveraged in a domain-agnostic model.
\begin{figure}
\vspace{-0.25cm}
  \centering
	\includegraphics[width=0.48\textwidth]{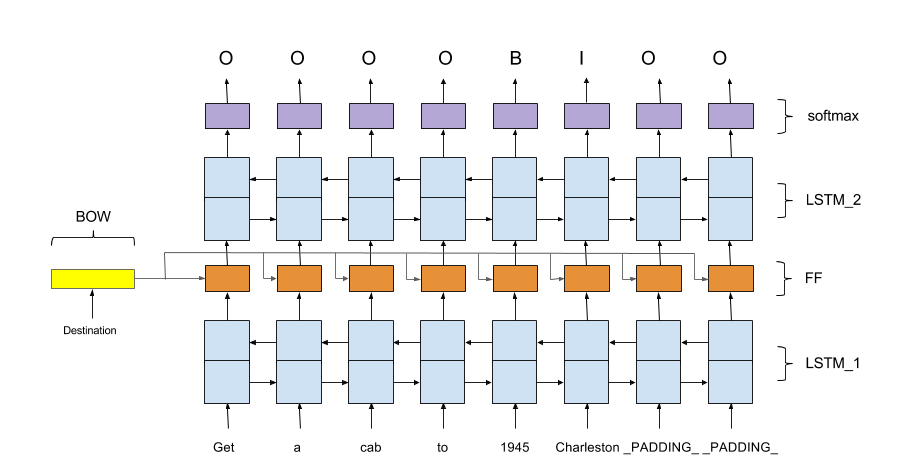}
	\caption{Zero shot Concept Tagger architecture}
	\label{fig:zsct}
\vspace{-0.5cm}
\end{figure}
An obvious example would be adding United Airlines when the multi-task
model can already parse queries for American Airlines and Turkish
Airlines. While the slot names may be different, the concept of {\em
  departure city} or {\em arrival city} should persist and can be
transferred to the new task of United Airlines using their natural language
descriptions. The very same idea can hold when the new domain is
related but not {\em flights}, but, say, {\em ground
  transportation}. Similarly, domain independent slots such as
location or date/time expressions can be implicitly shared for two
domains like {\em hotel reservation} or {\em restaurant reservation}. \\
In order to incorporate this slot description knowledge into model
training, we first generate an encoding of the slot by combining the token embeddings for the slot description. In principle this encoding can be obtained by passing description token embeddings through a RNN, but for the current experiments we just use their average. These slot representations are then combined within the multi-task architecture to obtain a domain-agnostic slot tagging model. \\
To elaborate, the zero-shot concept tagger consists of a single
256 dimensional bidirectional LSTM layer that acts on a sequence of tokens to produce contextual representations for each token in the utterance. This is followed by a feed forward layer where the
contextual token representations are combined with the slot encoding to produce vectors of 128 dimensions. This feeds into another 128 dimensional bi-directional LSTM layer followed by a softmax layer that outputs the prediction for that
slot. In our experiments these predictions are made independently for each slot by feeding a single slot description, but it is possible to slightly alter the architecture to make predictions for all slots. The input samples during training and evaluation for each slot included both positive (where the slot was present)
and negative samples (where it was absent). The ratio of training
samples from a particular domain in a batch was proportional to the
logarithm of the number of training samples.
The conceptual model architecture is depicted in
Figure~\ref{fig:zsct}.

\section{Experiments and Results}
\label{sec:exp}

\begin{table*}[t]
\centering
\caption{Weighted token F1 scores at various points on the learning curve for the compared models. ST corresponds to the single task baseline, MT corresponds to the multi task baseline and CT corresponds to the general concept tagging model.}
\label{tab:results}
\begin{tabular}{l | l | lll | lll | lll | lll | lll | lll |}
\hline
\# target train samples & 0    &     & 5    &     &   & 20   &   &  & 100  &  &  & 1000  & \\
\hline
Domain                  & CT   & ST   & MT   & CT   & ST   & MT   & CT   & ST   & MT   & CT   & ST   & MT   & CT\\
\hline
book\_room              & \bf{0.48} & 0.09 & 0.40 & \bf{0.49} & 0.28 & 0.53 & \bf{0.56} & 0.50 & 0.61 & \bf{0.65} & - & - & -\\
\hline
bus\_tickets            & \bf{0.45} & 0.11 & 0.40 & \bf{0.63} & 0.34 & 0.63 & \bf{0.72} & 0.60 & 0.74 & \bf{0.78} & - & - & -\\
\hline
flights\_1              & \bf{0.19} & 0.19 & 0.41 & \bf{0.42} & 0.46 & 0.59 & \bf{0.63} & 0.66 & 0.71 & \bf{0.75} & 0.77 & 0.79 & \bf{0.81}\\
\hline
flights\_2              & \bf{0.45} & 0.04 & 0.33 & \bf{0.56} & 0.33 & 0.59 & \bf{0.69} & 0.58 & 0.66 & \bf{0.73} & - & - & -\\
\hline
fare                    & \bf{0.04} & 0.53 & \bf{0.77} & 0.71 & 0.84 & 0.90 & \bf{0.92} & 0.93 & 0.94 & \bf{0.95} & \bf{0.98} & \bf{0.98} & \bf{0.98} &\\
\hline
book\_hotel             & \bf{0.45} & 0.08 & 0.31 & \bf{0.49} & 0.38 & 0.61 & \bf{0.64} & 0.69 & \bf{0.80} & 0.78 & 0.88 & \bf{0.90} & \bf{0.90}\\
\hline
find\_restaurants       & \bf{0.35} & 0.19 & 0.37 & \bf{0.52} & 0.44 & 0.60 & \bf{0.62} & 0.69 & \bf{0.75} & 0.74 & 0.82 & \bf{0.84} & \bf{0.84}\\
\hline
appointments            & \bf{0.56} & 0.24 & 0.55 & \bf{0.63} & 0.46 & 0.65 & \bf{0.67} & 0.65 & 0.72 & \bf{0.73} & 0.78 & 0.79 & \bf{0.80}\\
\hline
reserve\_restaurant     & \bf{0.56} & 0.20 & 0.50 & \bf{0.64} & 0.50 & 0.68 & \bf{0.70} & 0.72 & \bf{0.79} & 0.78 & 0.82 & 0.85 & \bf{0.86}\\
\hline
book\_cab               & \bf{0.18} & 0.46 & 0.63 & \bf{0.65} & 0.56 & 0.72 & \bf{0.74} & 0.78 & 0.82 & \bf{0.84} & 0.89 & 0.90 & \bf{0.92}\\
\hline
\end{tabular}
\end{table*}

We compare the performances of the models on varying amounts of
training data from each domain. This involves
using all available out of domain data and varying the amount of
training data for the target domain. To avoid
performance variations due to the small sample sizes, the performance was averaged over 10 runs with training samples drawn from different parts of the domain dataset. \\
For every domain, 20\% of the training examples were set aside for
the dev set. Since we were evaluating with varying amounts of training
data for each domain, the dev set was different for each data-point,
corresponding to the bottom 20\% of the training samples. For example,
if the training set consisted of 100 samples from Domain A and 20
samples from Domain B, dev set A would consist of 20 samples from
Domain A and dev set B would be comprised of 4 samples from Domain
B. The performance on these dev sets was evaluated separately and
averaged weighted by the log of the number of training samples. This
weighted average was used to tune the model hyper-parameters. We used
a logarithmic combination since it struck a good balance between noisy
evaluations on domains with small dev sets and over-tuning to the
domains with larger dev sets. \\
The performances along the learning curve for all the models on the 10 domains are described in Table~\ref{tab:results}. When no in-domain data is available, the concept tagging model is able to achieve reasonable bootstrap performance for most domains. Even when more data becomes available the model beats the single task model by significant margins and performs better than or on par with the multi-task baseline for most points on the learning curves.

\section{Discussion}
\label{sec:disc}
\begin{table}[]
\centering
\caption{Comparison of slot-wise performances of the concept tagging model (CT) and the multi-task model (MT) on ``appointment time" from appointments, ``pickup location" from book\_cab and ``\# seniors" from ``flights\_2".}
\label{tab:slotwise}
\begin{tabular}{l | l | l | l | l | l | l}
\# train samples & & 0    & 5            & 100           & 500           \\
\hline
\multirow{2}{*}{appointment time}       & CT & \textbf{0.84} & \textbf{0.85} & \textbf{0.88} &\textbf{0.89} \\\cline{2-6}
                 & MT & - & 0.66        & 0.87          & \textbf{0.89}          \\\hline
\multirow{2}{*}{pickup location}  & CT & 0.05 & 0.08          & 0.39          & 0.51      \\\cline{2-6}
                 & MT & - & \textbf{0.21} & \textbf{0.47} & \textbf{0.56} & \\\hline
\multirow{2}{*}{\# seniors}  & CT & \textbf{0.26} & \textbf{0.45}          & \textbf{0.70} & \textbf{0.75}      \\\cline{2-6}
                 & MT & - & 0.11 & 0.38 & 0.53 & \\\hline
\end{tabular}
\vspace{-0.35cm}
\end{table}
To better understand the strengths and weaknesses of the concept tagger we analyze its performance on individual slots. The model performs better than or on par with the multi-task model for most slots, with significant performance gains on slots that have shared semantics with slots in other domains. For slots that are specific to particular domains, like \emph{discount type} from \emph{bus\_tickets}, the concept tagger usually needs a larger number of training samples to reach the same level of performance. This can be explained by a lack of slot-specific parameters within the model. \\ Table~\ref{tab:slotwise} compares the performance of the concept tagger and the multi-task model on three slots across different domains that illustrate the strengths and weaknesses of our approach. By leveraging shared features and semantics with \emph{departure time}, \emph{reservation time} and time related slots from other domains the concept tagger is able to reach within 10\% of the peak performance on \emph{appointment time} without the need for any in-domain training data. Similarly, the model is able to ramp up performance on \emph{\# seniors} with a small amount of in-domain data, despite the presence of a competing slot with similar semantics (\emph{\# adults}) within the same domain.  This highlights the model's ability to generalize from slots with similar descriptions and semantics across domains.\\
On the other hand, the concept tagger's performance is worse than our multi-task baseline on \emph{pickup location}. A lack of a good contextual representations for the description \emph{pickup location} and the presence of a competing slot, \emph{dropoff location}, might be responsible for the performance degradation observed for this slot. This highlights the concept tagger's susceptibility to descriptions that fail to produce a compatible slot representation, either due to an incomplete or misleading description of the slot semantics or a lack of good embedding representations for these descriptions. It might be possible to alleviate poor slot representations by fine tuning the slot representations on small amounts of in-domain training data after starting with representations derived from pre-trained word embeddings or using contextual word embeddings~\cite{huang2012improving}. Enhancing utterance token representations with an entity linker or a knowledge base are possible extensions of this work that might enable better generalization to new entities. Exploring use of unlabeled training data from target domains with a domain adversarial loss~\cite{ganin2015unsupervised} might be another interesting avenue for exploration.\\
The appeal for our approach derives from its simplicity and the minimal amount of supervision required to bring up slot-filling on a new domain. The proposed solution makes it possible to design a plug and play system with a reduced need for expensive labeled data for every additional domain.

\section{Conclusions}
\label{sec:conc}
In this paper we propose a novel approach to slot filling that leverages shared feature extraction and slot representations across domains by using the natural language descriptions of slots.
We crowd-sourced slot filling datasets for ten domains to explore approaches that can easily scale across domains and demonstrate that our proposed concept tagging model performs significantly better than a strong multi-task baseline, especially in the low data regime.
To further evaluate the strengths and weaknesses of the proposed approach, we analyze its performance on individual slots and demonstrate that our model is able to leverage shared features and semantic descriptions of slots defined in other domains, and shows potential for reasonable performance on slots in a new domain without the need for any in-domain training data or explicit schema alignment. \\ 
We hope that our proposed solution can provide a baseline approach for future research into scalable frame semantic parsing systems.

\bibliographystyle{IEEEtran}
\bibliography{all}{}

\end{document}